\documentclass[man,floatsintext]{apa7}

\usepackage{csquotes}
\usepackage{multirow}
\usepackage{multicol}
\usepackage{graphicx} 
\usepackage{rotating} 
\usepackage{graphicx}
\usepackage{subcaption}
\usepackage{tabularx}
\usepackage[style=apa, backend=biber]{biblatex}
\usepackage{hyperref}  
\usepackage{placeins}
\usepackage{setspace}

\hypersetup{breaklinks=true} 

\newcommand{\templategap}[1][1.0cm]{\underline{\hspace{#1}}}

\title{Automatic Detection of Inauthentic Templated Responses \\in English Language Assessments}
\shorttitle{Automatic Detection of Inauthentic Templated Responses}
\author{Yashad Samant\textsuperscript{1}, Lee Becker\textsuperscript{2}, Scott Hellman\textsuperscript{2}, Bradley Behan\textsuperscript{2}, Sarah Hughes\textsuperscript{2\textdagger}, Joshua Southerland\textsuperscript{2\textdaggerdbl}}
\affiliation{
    \textsuperscript{1} Research conducted while at Pearson Education\\
    \textsuperscript{2} {Pearson Education, Inc.}
}
\authornote{
    Correspondence concerning this article should be addressed to Lee Becker. \\
    Pearson affiliated authors can be reached at \texttt{<given>.<surname>@pearson.com}.  \\
    \textsuperscript{\textdagger}Sarah Hughes can be reached at sarah.hughes1@pearson.com. \\
    \textsuperscript{\textdaggerdbl}Joshua Southerland can be reached at josh.southerland@pearson.com
}
\date{March 2025}

\begin{document}
\maketitle

\begin{abstract}
. In high-stakes English Language Assessments, low-skill test takers may employ memorized materials called ``templates'' on essay questions to ``game'' or fool the automated scoring system. In this study, we introduce the automated detection of inauthentic, templated responses (AuDITR) task, describe a machine learning-based approach to this task and illustrate the importance of regularly updating these models in production. 
\end{abstract}

\section{Introduction}

English language proficiency (ELP) tests carry exceptionally high stakes because of how they influence access to employment, education and national residency status.  ELP test authors use constructed response essay questions because they offer deeper understanding of the test taker's language skill compared to multiple-choice or other selected-response item formats.  By requiring test takers to write their own response, they can assess depth of understanding, authentic language use, fluency and communication skill. At the same time, it is widely known that major test administrators utilize automated essay scoring (AES) technologies to enable assessment at scale.  This convergence of high stakes and automation can incentivize test takers to seek mechanisms to ``game'' or fool scoring systems into assigning higher scores than warranted. 

A straightforward gaming strategy is to memorize and respond to questions with an essay known to yield a high score.  However, this approach becomes prohibitive when a test draws from a large pool of items. In recent years, test takers have turned to a strategy centered on templates -- fixed texts with slots for customization.  This approach eliminates the need to memorize a text per item, and it also grants test takers the ability to customize their response to a large number of items.
  Early successes with this tactic have led to a proliferation of templates on the Internet. An online search for ``PTE templates'' will yield a wide range of forums, social media influencers and test preparation companies that promote their templates as a sure-fire way to ensure a high score. 

Figure~\ref{fig:example-template} shows an example template found via web search.  Like a high-scoring essay, a template aims to include the expected elements of effective writing such as advanced vocabulary, regular use of discourse markers, a variety of sentence constructions and minimal grammar and mechanics errors.  Unlike a fixed essay, templates also include gaps that allow the test taker to customize their response to the specific essay prompt. 

\begin{figure}[b!]
    \caption{\textit{Example essay template}}
    \begin{center}
        \parbox{0.85\linewidth}{
            \begin{quote}
            \footnotesize 
            \singlespacing One of the most conspicuous trends of today's world is a colossal upsurge in the number of people believing that \templategap.  There is a widespread worry that this only lead to a myriad of concerns in one's life. However, I do not entirely agree with this and I will explain why in this essay. There are a number of arguments in favour of my stance. The most preponderant one is that \templategap can not only contribute to \templategap, but also lead to numerous other benefits in various fields. Thanks to the wide range of advantages it offers, not only can one benefit more when it comes to being effective, but they can also enhance the productivity and quality of their lives, with much ease, efficacy and convenience. Needless to say, all these merits stand one in good stead, as far as augmenting their chances of prosperity and excellence is concerned. Another pivotal aspect of the aforementioned proposition is that it is only likely to help one thrive and excel in varied areas. Besides, only when one follows such a system, can they broaden their horizons, thus learning such an essential attributes as responsibility, dedication and perseverance. Hence, it is apparent why many are in favour of the topic. In view of the arguments outlined above, one can conclude that the benefits of the \templategap are indeed too great to ignore.
            \end{quote}
        }
    \end{center}
    \par \small Note. Template retrieved from \url{https://www.coursehero.com/file/plrhnk/One-of-the-most-conspicuous-trends-of-todays-world-is-a-colossal-upsurge-in-the/}.

    \label{fig:example-template}
\end{figure}

The remainder of this paper will give a background on related research in gaming the system, describe the data collection and machine learning experiments needed to train an operational AuDITR model, discuss operational concerns of such a model and close with discussion and future research directions.

\section{Related Works}

\textcite{hughes_maintaining_2024} defines \emph{gaming} as ``the use of construct-irrelevant response strategies that misrepresent
or obscure a test taker’s true ability.''  Within the context of AES, a key motivation for gaming the system is to obtain a higher score from a scoring system than warranted \parencite{lottridge_comparing_2020}.

There is a broad body of research around techniques for gaming AES systems and for detecting gaming behaviors. Common techniques include duplication and repetition of text \parencite{higgins_managing_2014, zhang_evaluating_2016}; generation of gibberish \parencite{perelman_babel_2020, cahill_developing_2018}; and injection of topical keywords \parencite{higgins_managing_2014, zhang_evaluating_2016}.  Off-topic essay detection \parencite{higgins_identifying_2006, burkhardt_examining_2013, higgins_managing_2014, hendersen_gamification_2019} is a gaming adjacent task, as test takers may attempt to respond with non-prompt related essays or text. However off-topic is not strictly gaming as off-topic responses may also occur from test takers misunderstanding the question, entering their response in the wrong box or a desire to cause mischief .

While previous works have discussed the prevalence of templates for ELP assessments \parencite{zou_case_2024, hughes_maintaining_2024, kim_online_2023, kim_prepping_2021, xi_use_2024}, to the best of our knowledge, there is no literature regarding the detection of templated responses in ELP assessments.

\section{Automated Detection of Templated Gaming}

\subsection{Corpus and Annotation}

To enable training and evaluation of machine learning-based approaches to detecting templated responses, we developed a corpus of responses to essay items from the Pearson Test of English.   Our initial data exploration and pilot labeling suggested that forcing a binary distinction between templated and non-templated behaviors makes the consistent labeling difficult.  In some cases, test takers started with memorized templates, but also contributed a large amount of their own authentic writing.  Because our end goal is to have models capable of being tuned for a variety of operating points, we opted to frame the annotation task as a multi-label classification / scoring scheme.  The scores and labels are defined in Table~\ref{tab:response-labeling-schema}.

For the modeling experiments, we assembled a corpus of 319 responses from multiple essay items.  These essays were labeled by subject matter experts (SMEs) using the schema in Table~\ref{tab:response-labeling-schema}.  Unfortunately, we no longer have inter-rater reliability metrics to report for ternary labeled data (see the Limitations section).  Instead Table~\ref{tab:oscar4k-inter-rater-reliability} contains a confusion matrix for a different annotation effort using binary labels (0-no templating, 1-templating detected). In the original ternary data, agreement was strongest for responses which display significant templating and for those that exhibit no gaming.  In the binary data, the 77.25\% exact agreement and Kappa of 0.446 also reflect the difficulty of assigning labels which exhibit some degree of templating as each rater may have a different notion of penalization.

Recognition and recall of templates is another key challenge in labeling these data.  Often, the first time a rater sees a template, they give the test taker the benefit of the doubt.  It is only after seeing the nearly identical language multiple times that a rater realizes that the current and previous responses should all be labeled with a \textit{Low} or \textit{High} label.  Because raters score in a streaming fashion, there is some inherent noise in the labels.


\begin{table}[!t]
    \centering
    \caption{Response Labeling Schema}
    \begin{tabular}{c|l|l}
         \textbf{Score point} & \textbf{Label} & \textbf{Description} \\
         \hline 
         0 & None & Little or no templating detected \\
         1 & Low & Some templating detected \\
         2 & High & Significant templating detected
    \end{tabular}
    \label{tab:response-labeling-schema}
\end{table}


\begin{table}[!t]
    \caption{Binary Label Inter-Rater Reliability}

            \begin{center}
                \begin{tabular}{c|c|c|c|c|}
                    \multicolumn{2}{l}{} & \multicolumn{3}{c}{\emph{Second Read}} \\
                    \cline{3-5}
                    \multicolumn{2}{c|}{} & \textbf{0} & \textbf{1} & \textbf{Total} \\
                    \cline{2-5}
                    \multirow{3}{*}{\rotatebox{90}{\emph{First Read}}}
                    & \textbf{0} & 698 & 443 & 1141 \\
                    \cline{2-5}
                    & \textbf{1} & 467 & 2392 & 2859\\
                    \cline{2-5}
                    & \textbf{Total} & 1165 & 2835 &  4000 \\
                    \cline{2-5}
                \end{tabular}
            \end{center}

        \hfill
        \label{tab:oscar4k-inter-rater-reliability}
\end{table}


\subsection{Model}

Inauthentic, templated-response detection models do not run in isolation.  They are paired with the automated scoring model to ensure that responses do not receive scores higher than warranted.  This automatic setting favors operating points with high precision to reduce the likelihood of false positives.  Consequently any model deployed to production must allow for effective calibration with thresholds, enabling us to adjust the sensitivity to maintain high precision while ensuring adequate recall.

At first glance, detection of templated responses looks like a standard document classification task. The occurrence of n-grams from templates in a response is a strong indicator that a template was used.  However, the current default of fine-tuning a Transformer \parencite{vaswani_attention_2017} like BERT \parencite{devlin_bert_2019} or training a model with features like TF-IDF or n-gram counts is prone to overfitting to the templates seen in the training data.  To allow for more better recall on unseen responses,
we engineer features which are interpretable and aligned to the core determination of template usage.

While in operation, we may learn of new templates by automatic discovery processes, mining of web and social media, or curation by human scorers.  To allow for more timely updates, we prefer a model that allows update via template configuration over additional retraining.

To best support these requirements, we frame this task as the binary classification problem:  determining whether a response exhibits \textit{High} gaming (labeled 1/True) or \textit{Low} or \textit{No} gaming (labeled 0/False). Model inputs consist of the response text, prompt text and a list of templates.

\subsection{Features}

When we talk broadly about the AuDITR task, we consider as \textit{authentic} texts those that the test taker wrote on their own, and \textit{inauthentic} texts as those that they produced from other resources with little use of their own skill.  Our feature space quantifies regions of the text that overlap with templates and prompt texts.  To provide more granularity in matching, we segment full templates into smaller regions called sub-templates.  This has an additional benefit of detecting templated text when test takers mix and match texts from multiple templates.

We use a sliding n-gram window over the response texts and compare against n-gram windows over the prompt text and sub-template texts with an efficient Levenshtein distance match \parencite{levenshtein1966binary}.  This match process identifies regions of the text that overlap with the input prompt text and templates.  This match data is then used to calculate the summary features defined in Table~\ref{tab:auditr-model-features}.

\begin{table}[t!]
    \caption{AuDITR Model Features}
    \centering
    \small
        \begin{tabularx}{\linewidth}{|p{4.5cm}|X|}
        \hline
        Feature &  Description \\
        \hline
        num-non-template-tokens & Number of tokens in response that do not match a sub-template region. \\
        \hline
        pct-non-template-tokens & Percent of tokens in response that do not match a sub-template region. \\
        \hline
        num-non-prompt-tokens & Number of tokens in response that do not match a prompt text substring. \\
        \hline
        pct-non-prompt-tokens & Percent of tokens in response that do not match a prompt text substring. \\
        \hline
        num-authentic-tokens & Number of tokens in response that neither match a sub-template region nor a prompt text substring \\
        \hline
        pct-authentic-tokens & Percent of tokens in response that neither match a sub-template region nor a prompt text substring \\
        \hline
    \end{tabularx}
    \label{tab:auditr-model-features}
\end{table}

\subsection{Experiments and Results}

The corpus of 319 responses was partitioned into a training set with 213 responses and a test set with 106 responses.  As stated above, the \textit{None} and \textit{Low} label responses were combined into a 0 label and the \textit{High} label responses were given a 1 label.  We use a random forest classifier with grid search over the max depth, max features, and number of trees hyperparameters.  Tables~\ref{tab:train-set-agreement} and \ref{tab:test-set-agreement} contain confusion matrices that break down agreement by label; Table~\ref{tab:performance-metrics} summarizes overall performance in terms of Precision ($P$), Recall and $F_{1}$ scores.

While there is a small drop in recall and $F_1$ between the training and test sets, the absence of false positives in the test set suggest that the model can add benefit even at a conservative operating point.  An error analysis of the false positives from the training set, found that all four responses contained templated language.  Conversely analysis of false negatives found responses with some templated language, but not to the degree we see in ones where the model is confident.  In these cases the human raters were more aggressive in detection than the model.  In other cases, the responses contained evidence of templated language that was not part of the match list used as input.

\begin{table}[!t]
    \caption{Model Performance}
        \begin{subtable}[b]{0.30\textwidth}
            \small
            \begin{center}
                \begin{tabular}{c|c|c|c|c|}
                    \multicolumn{2}{l}{} & \multicolumn{3}{c}{\emph{Model}} \\
                    \cline{3-5}
                    \multicolumn{2}{c|}{} & \textbf{0} & \textbf{1} & \textbf{Total} \\
                    \cline{2-5}
                    \multirow{3}{*}{\rotatebox{90}{\emph{Human}}}
                    & \textbf{0} & 99 & 4 & 103 \\
                    \cline{2-5}
                    & \textbf{1} & 32 & 78 & 110 \\
                    \cline{2-5}
                    & \textbf{Total} & 131 & 82 & 213 \\
                    \cline{2-5}
                \end{tabular}
                \caption{Training Set Agreement}
                \label{tab:train-set-agreement}
            \end{center}
        \end{subtable}
        \hfill
        \begin{subtable}[b]{0.30\textwidth}
            \small
            \begin{center}
                \begin{tabular}{c|c|c|c|c|}
                    \multicolumn{2}{l}{} & \multicolumn{3}{c}{\emph{Model}} \\
                    \cline{3-5}
                    \multicolumn{2}{c|}{} & \textbf{0} & \textbf{1} & \textbf{Total} \\
                    \cline{2-5}
                    \multirow{3}{*}{\rotatebox{90}{\emph{Human}}}
                    & \textbf{0} & 51 & - & 51 \\
                    \cline{2-5}
                    & \textbf{1} & 23 & 32 & 55 \\
                    \cline{2-5}
                    & \textbf{Total} & 74 & 32 & 106 \\
                    \cline{2-5}
                \end{tabular}
                \caption{Test Set Agreement}
                \label{tab:test-set-agreement}
            \end{center}
        \end{subtable}
        \hfill
        \begin{subtable}[b]{0.30\textwidth}
            \begin{center}
                \begin{tabular}{|c|c|c|c|}
                    \hline
                    Split & $P$ & $R$ & $F_{1}$ \\
                    \hline
                    Train & 95.1 & 70.9 & 81.2 \\
                    Test  & 100  & 58.2 & 73.6 \\
                    \hline
                \end{tabular}
                \caption{Performance Metrics}
                \label{tab:performance-metrics}
            \end{center}
        \end{subtable}
    \par \small Note.  In agreement confusion matrices, rows represent the label assigned by humans.  Columns represent the label assigned by the model.  Results come from a default classifier threshold of $0.8$.
    \label{fig:model-performance}
\end{table}

\section{Operating Concerns}

Beyond model training, additional calibration is required to pick a production operating point.  In practice another corpus of recent operationally-scored responses is used to determine detection rates at various thresholds, and we hand-label a sample of the gaming-detected responses to ensure no test taker is unfairly penalized.

After the release of the first AuDITR model, we observed a steady decline in our model's detection rate, leading us to hypothesize that there was a shift in test taker response behaviors.  We doubted that test takers had fully abandoned the use of templates, but rather had switched to a new set of templates.  Inspection of recent responses and monitoring of social media  confirmed our suspicions.  Influencers and discussion forums spoke of how the ``algorithm had changed'' and ``PTE banned templates''.

To address the degradation in detection rates, we not only updated our template list with templates found online, but also developed new processes to automatically discover new templates and sub-templates.  A description of the template discovery process is outside of the scope of this paper, so we will focus only on the effects of expanded template coverage and model updates.

\begin{figure}
    \caption{AuDITR detection rates over time}
    \centering
    \includegraphics[width=0.75\linewidth]{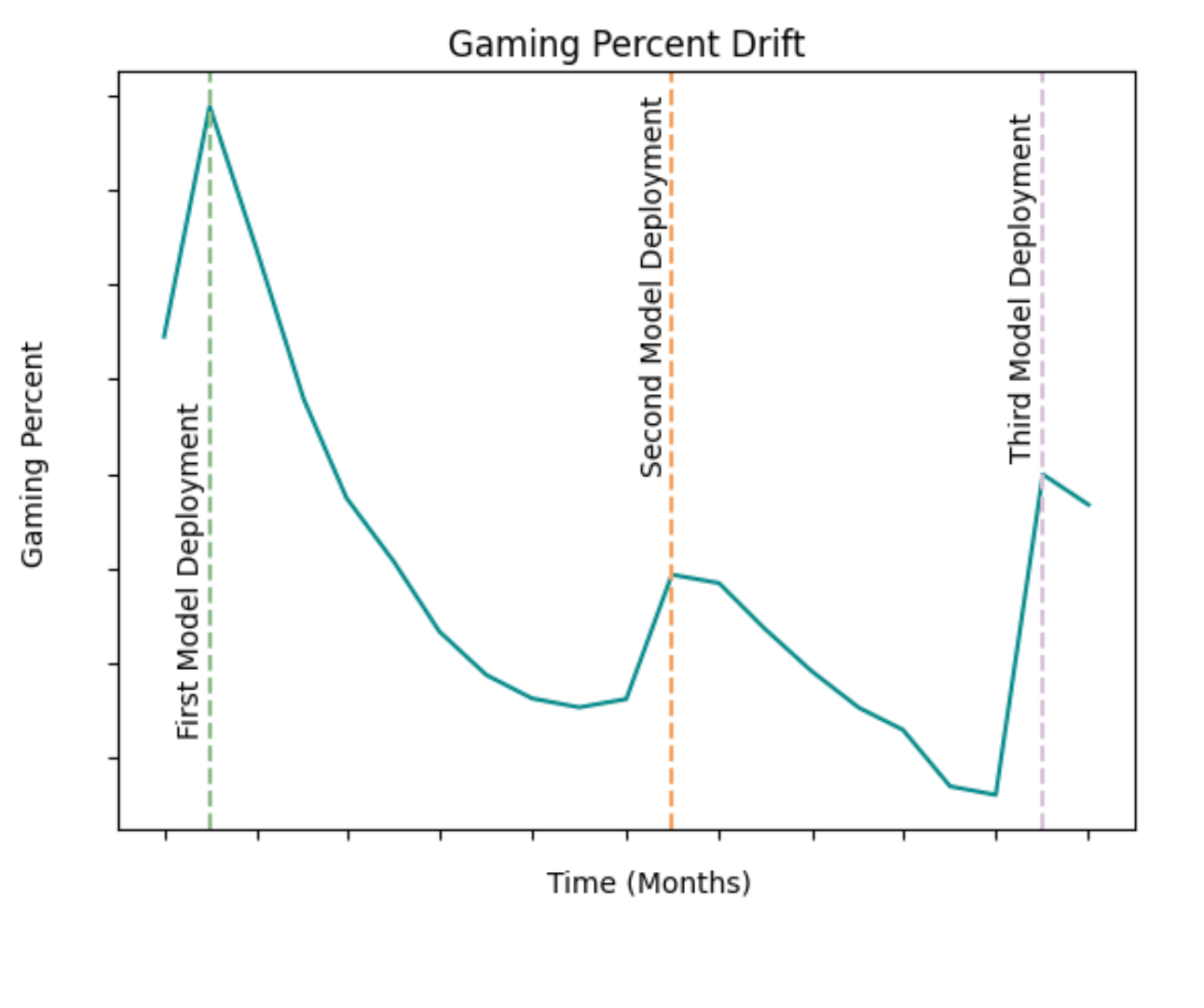}
    \label{fig:model-detection-curve}
    \par \small Note. The actual percentages and dates have been redacted to protect proprietary information.
\end{figure}

Figure~\ref{fig:model-detection-curve} shows gaming percent drift over the course of several months.  As indicated by the vertical dotted lines, each model release brings a spike in detection rates.  As word gets out that the ``algorithm'' has changed, the prescribed templates and strategies shift, which in turn brings about a decay in detection rates until the next model is released. This dynamic known as adaptive adversarial drift has parallels in other domains including spam detection, cybersecurity and fraud detection where the introduction of a model causes bad-faith actors to find workarounds \parencite{dalvi2004adversarial}.

\section{Discussion and Future Work}

The cat-and-mouse tension between model release and test takers adopting new templates represents a unique challenge for English Proficiency Testing.  Due to the high-stakes nature of ELP assessments, many test takers of the PTE and other English Language exams are motivated to achieve high scores via any advantage possible.  Whether they are taking the test multiple times or turning to a global network of influencers and businesses promoting their tips and tricks, the result is a dynamic ecosystem where today's ability to detect templated response is not relevant tomorrow.

That said, we are also encouraged by the shift in advice we have observed since releasing our first models.  Online videos increasingly remind test takers that a template alone will not achieve a high score and that to get full credit they must write topically to the prompt, insert more than single words into gaps, and ensure the text they add is free of grammar, usage, mechanics and spelling errors.  In other words, to get a high score, test takers should address the question and write correct English.

Improvement of our AuDITR model is an on-going area of research.  For future experiments, we plan to investigate how techniques from forensic linguistics, author identification and plagiarism detection can tease apart authentic and inauthentic regions of the text.  We also believe there is promise in utilizing generative language models to aid in learning deep representations of templated behavior that goes beyond our existing matching mechanisms.  Addressing model drift via adversarial classification  \parencite{dalvi2004adversarial} and drift sensitive learners \parencite{grosshans2013detection} presents another area for investigation and these approaches can reduce the need for ad-hoc training and can create models that are aware of when drift occurs.

Lastly, our human scoring operations plan to move away from binary gaming labels to the ternary ones.  This change will give us a larger, more reliable stream of data for training, evaluation and update of models.

\section{Limitations}\label{limitations}

One limitation of our study is the loss of the original dataset with ternary labels due to company data retention policies. As a result, we were unable to report inter-rater reliability for the ternary labels. We acknowledge that this impacts the ability to fully validate the consistency of the labeling process. To mitigate this, we used a derived dataset with binary labels and a single annotation for training our model.

\printbibliography

\end{document}